\documentclass[final]{cvpr}
\usepackage{times}
\usepackage{epsfig}
\usepackage{graphicx}
\usepackage{amsmath}
\usepackage{amssymb}
\usepackage{subcaption}
\usepackage{array}
\usepackage{multirow}
\usepackage{gensymb}
\usepackage{microtype}
\usepackage{booktabs} %
\usepackage{enumitem}
\usepackage{booktabs} %

\usepackage[utf8]{inputenc}
\usepackage[font=small,labelfont=bf]{caption}

\DeclareMathOperator*{\argmax}{arg\,max}

\usepackage[pagebackref=true,breaklinks=true,colorlinks,bookmarks=false]{hyperref}

\begin{document}

\title{Discovering Underground Maps from Fashion}

\author{Utkarsh Mall\textsuperscript{1, 2}\qquad
Kavita Bala\textsuperscript{1} \qquad Tamara Berg\textsuperscript{3} \qquad  Kristen Grauman\textsuperscript{2,4} \\
\textsuperscript{1}Cornell University, \textsuperscript{2}Facebook AI Research, \textsuperscript{3}Facebook,
\textsuperscript{4}University of Texas at Austin\\
{\tt\small {utkarshm@cs.cornell.edu}} \qquad {\tt\small {kb@cs.cornell.edu}} \qquad {\tt\small {tlberg@fb.com}} \qquad {\tt\small {grauman@cs.utexas.edu}}
}

\maketitle

\begin{abstract}
The fashion sense---meaning the clothing styles people wear---in a geographical region can reveal information about that region.  For example, it can reflect the kind of activities people do there, 
or the type of crowds that frequently visit the region (e.g., tourist hot spot, student neighborhood, business center).
We propose a method to automatically create \emph{underground neighborhood maps} of cities by analyzing how people dress. %
Using publicly available images from across a city, our method finds
neighborhoods with a similar fashion sense and segments the map
without supervision. For 37 cities worldwide, we show promising
results in creating good underground maps, as evaluated using
experiments with human judges and underground map benchmarks derived
from non-image data. Our approach further allows detecting distinct
neighborhoods (what is the most unique region of LA?) and answering analogy questions between cities (what is the ``Downtown LA" of Bogota?).
The supplementary can be found at: \href{http://www.cs.cornell.edu/~utkarshm/underground_maps/supplementary.pdf}{www.cs.cornell. edu/$\sim$utkarshm/underground\_maps/supplementary.pdf}
\end{abstract}

\emph{``The map is not the thing mapped."}---Eric Temple Bell

\section{Introduction} \label{sec:intro}
\begin{figure}
\centering
\includegraphics[width=\linewidth]{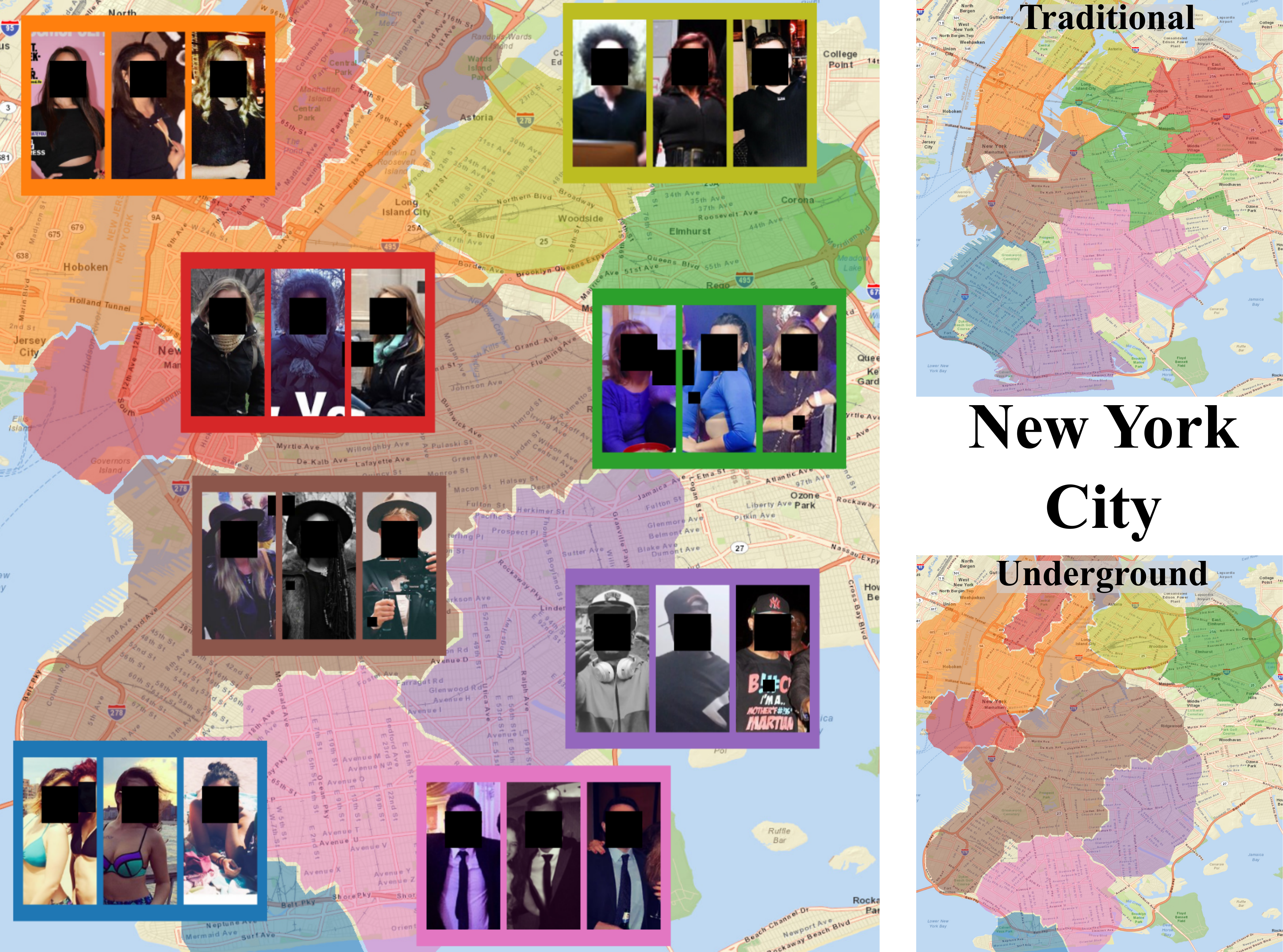}
\vspace*{-0.3in}
\caption{\small Discovered underground map of New York City with 8 neighborhoods.  \textbf{Left:}  The photos show the top discriminative styles from each neighborhood corresponding to the color of their border.
\small \textbf{Right:} Map produced from administrative boundaries (``Traditional'', top) vs.~the map produced by our method (``Underground'', bottom). Traditional neighborhoods do not capture information made apparent by an underground map, e.g., discovering two far apart but similar tourist neighborhoods (red).}
\label{fig:fig1}
\end{figure}

Cities are traditionally divided into multiple neighborhoods, where neighborhood boundaries occur due to a variety of reasons, including city governance and management, geographic separation of regions (e.g., by water, hills, etc.), or historical factors (past city extension). 
However, a person who knows a city well often has a different notion of neighborhoods than what these boundaries provide.  
To take New York City as an example, even though Manhattan is divided into Downtown, Midtown, and Uptown by traditional maps, a local person may see a region in Uptown (Columbia University) as similar to a region in Downtown (NYU), since student populations live in both regions. 
Similarly, there are tourists south of downtown and throughout most of midtown. Lower Manhattan is not a monolithic neighborhood, but rather a culturally diverse region, e.g., encompassing both Chinatown and corporate offices near Wall Street.
Often such notions of neighborhoods can be more useful than  traditional neighborhood divisions, as  they reveal how a neighborhood actually is perceived and experienced. 
We call these kind of neighborhood maps ``underground'' maps to differentiate them from  traditional neighborhood maps.

How can we get such underground maps? 
While no prior work infers underground maps, past computer vision work explores ``urban perception" using street-view images of buildings~\cite{Ordonez-14, khosla-14, dubey-16} and cars~\cite{Gebru-17} to characterize a location.  
While fascinating, such glimpses of a city remain a step removed from the people who traverse it, and they are static (e.g., buildings may persist unchanged for decades, while the culture of a neighborhood evolves more rapidly).
Meanwhile, directly crowdsourcing for an underground map is challenging to scale and requires city-specific expertise.

We propose to discover underground neighborhood maps from \emph{fashion senses} observed in public social media photos. See Figure~\ref{fig:fig1}.  
The key insight is that people's clothing is a strong indicator of their personal style,  interests, and current activity, which in turn reveals a bottom-up grouping of the regions within a city.  
For example, people in the vicinity of a beach are likely to be found wearing beachwear, whereas people preparing to jog may wear short-sleeved shirts and shorts in warm weather. 
Similarly, students near a university often wear shirts with their university colors, while sports fans don the colors of their team, and others wear clothes reflecting a social cause or pop culture element that may be active in a part of a city.
Unlike architecture or cars, fashion images provide dynamic information.
For example, a person drives the same car in the whole city irrespective of what they plan to do in a particular neighborhood, whereas their clothes can change based on the activity (\textit{e.g.,} gym vs. beach vs. work).
Based on these observations, we believe that fashion is an interesting yet unexplored indicator of the underground map notion. 

Our approach uses the distribution of fashion styles (we call this {\em fashion sense}) at a place to discover an underground map. 
We first detect clothing attributes in 7.7M public geo-located social media photos spanning 37 cities worldwide, and then discover typical combinations of those attributes, or styles.  
Then, we use unsupervised clustering to detect pockets of a city that are both spatially and stylistically coherent.  
Finally, in addition to returning the generated neighborhood map, we devise computational measures to identify a city's most unique neighborhoods and mine for ``analogical" neighborhoods between otherwise different cities (e.g., what is the ``Uptown" of Bogota?).  
In contrast to previous work on urban perception, which requires supervision in the form of image labels~\cite{Ordonez-14,khosla-14,dubey-16,Murdock-15} our method uses no underground neighborhood labels. 
Figure \ref{fig:fig1} shows an example underground map created by our method for New York City, compared to a traditional administrative city map (cf.~Sec~\ref{sec:results}).

There are many potential applications of underground maps.
A person unfamiliar with a city could find out what neighborhoods might be suitable for them to visit, e.g.,  to satisfy interests in outdoor activities vs.~shopping vs.~tourist areas.
A visitor could grasp at a glance how people typically dress in a region, e.g., when choosing attire for a restaurant. %
Anthropologists could leverage the mined maps to infer trends within a city and across time. A more obscure part of a city could gain positive exposure for its distinct culture.
In any such case, an underground map addresses queries in ways that go beyond traditional maps.

We evaluate our method quantitatively on two non-visual benchmarks that capture  notions of underground maps. 
One benchmark captures how people perceive a neighborhood, while the
other captures the business distribution (indirectly the activity distribution) in a neighborhood.
Experiments %
show that our model is able to produce accurate and coherent neighborhood regions.
Further, our qualitative results and evaluation with human judges reinforce these findings and illustrate the value of fashion images as a new tool to interpret subtleties in the life of a city.
Our work is the first to discover unsupervised underground maps of a city and to analyze fashion \emph{within} individual cities at a large scale.

\section{Related work} \label{sec:relwork}

\textbf{Visual understanding of clothing.} 
Computer vision techniques are actively used for fashion.
Research has focused on the classification of clothing attributes \cite{Chen-12,Bourdev-11,Bossard-12,Zhang-14, Liu-16, matzen-17},
segmenting clothing in images~\cite{Yamaguchi-12,Yamaguchi-13,Yang-14, jia-20}, and product identification to  retrieve specific clothing products in photos~\cite{Di-13,Vittayakorn-15,Kiapour-15, Liu-16}.  
There is also prior work on classifying the ensemble of clothing a person is wearing (style), e.g., ``hipster'', ``goth''~\cite{Kiapour-14}.
Clothing recommendation systems build models for various requirements like occasion-based dressing~\cite{liu-12b}, occasion and location~\cite{ma-19}, or %
compatibility and style coverage~\cite{hsiao-18}. 
Our work uses attribute prediction to create an embedding space for understanding the fashion sense of a city. However, the goal is not just to classify fashion attributes on images, but to use the embedding to discover underground neighborhood maps for a city. 

\textbf{Visual style discovery.}
Some prior research uses visual analysis to discover styles and trends.
Early work used low-level image features or mined visually distinctive patches~\cite{Doersch-12, Singh-12, Doersch-13} to discover unique architectural properties of a city without supervision.
Discovering fashion styles without style supervision has also been explored~\cite{matzen-17,Hsiao-17,AlHalah-17}. These methods leverage attribute predictions or embeddings %
to discover distinct looking styles. Learning to detect urban tribes by using crowdsourced data %
has also been explored~\cite{kwak-13}.  
While these methods focus on discovering different styles, we leverage styles to discover neighborhoods with different fashion senses.

\textbf{Trend forecasting.}
Recent work in fashion trend forecasting trains temporal models to predict how a particular fashion style will rise or fall in popularity in the future. 
This was first addressed in \cite{AlHalah-17}, who make quantitative forecasts of fashion trends with coarse yearly predictions for a year in advance. Recent work looks at finer-grained trend forecasting at a weekly granularity~\cite{mall-19, ma-20, alhalah-20}. %
These models enable discovery of unique events where people wear a specific type of clothing with an anomalously high occurrence~\cite{mall-19} and  detection of fashion influence of one city on another~\cite{alhalah-20}. 
Unlike existing methods, our work aims to discover latent maps of cities using the fashion sense of regions \emph{within} a city.
This is a  challenging task as we do not know the boundaries of neighborhoods a priori; our goal is to discover them without any supervision.
We are the first to discover latent neighborhoods using fashion sense. %

\textbf{Urban perception.} 
Prior work has focused on predicting geo-spatial properties within an urban environment, using the visual features at a geo-location. This includes properties like the perceived safety of cities~\cite{Arietta-14,Naik-14,Ordonez-14, dubey-16}, ecological properties such as snow or cloud cover~\cite{Zhang-12,Wang-13,Murdock-15}, or the distance to Starbucks~\cite{khosla-14}. 
Advances in visual recognition have enabled more sophisticated analyses, such as modeling demographics by recognizing the make and model of cars in StreetView~\cite{Gebru-17} or by recognizing structures in satellite images~\cite{perez-17}.
Note that all these previous methods require some form of labeling for the latent attribute they are trying to perceive. For example, human annotators manually rank images %
based on safety~\cite{Ordonez-14}, or 
demographic data of poverty is needed to predict poverty in an unseen region~\cite{perez-17}. In contrast, our approach does not use any such labels for the segmentations it produces. To our knowledge, we are the first to perceive such a factor without using any supervision, and the first to address the problem of underground maps.

\section{Method} \label{sec:method}

\begin{figure}[t]
\centering
\includegraphics[width=\linewidth]{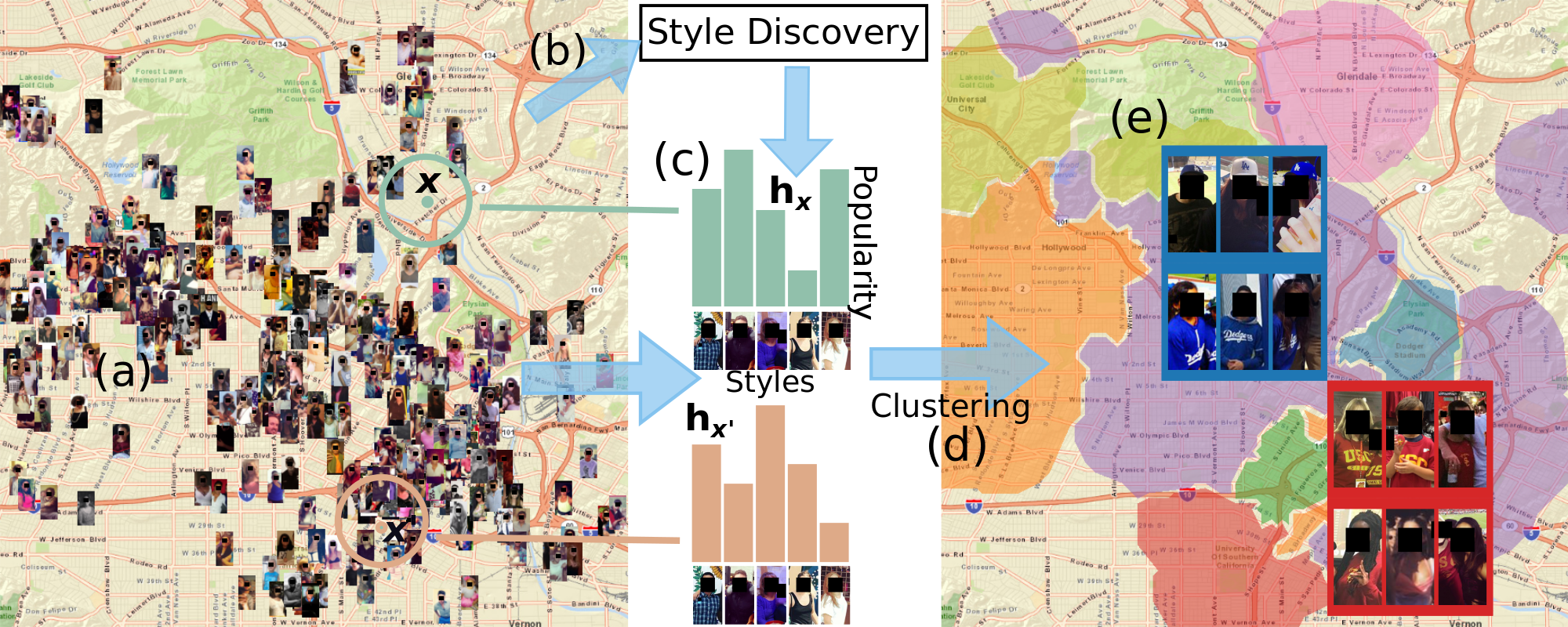}
\vspace*{-0.2in}
\caption{\small Pipeline to discover fashion neighborhoods. (a) Input: large set of geo-tagged fashion images. (b) We perform unsupervised style discovery to get styles. (c) Using the styles and geo-tagged images, we describe a location $x$ using its style histogram $h_x$. (d) Finally, we cluster locations using the descriptors. (e) Neighborhoods can be visualized by looking at their top styles. %
}\label{fig:pipeline}
\end{figure}

Our goal is to segment the map of a city into regions based on the fashion sense in the region. To do this, we need a model that can understand fashion elements in an image. We use this model with our framework to segment a city into underground neighborhoods. %
Next we provide background on the problem and the recognition model we use, and we then discuss our method. Fig. \ref{fig:pipeline} overviews our method.

\subsection{Dataset and Style Discovery} \label{ssec:back}
To understand fashion sense within a neighborhood of a city, we aim to understand the clothing in images of people in that neighborhood.
Therefore, we need a dataset that can capture how people dress at  different locations in a city.
To get a real-world unfiltered glimpse of what people are wearing across a city, we use images sampled from social media platforms; specifically, we 
use the 7.7M images from the GeoStyle dataset \cite{mall-19} from Instagram and Flickr.

For a city, let $\{I_i\}$ be the set of images of people.  
We also know the geolocation tagged to each of these images, $\boldsymbol{l_i} \in \mathbb{R}^2$ for an image $I_i$.
Following the method in \cite{mall-19}, we first train a representation by learning to classify basic clothing attributes. 
The attributes consist of a variety of fashion properties, like clothing type (suit, t-shirt, dress etc.), presence of accessories (sunglasses, necktie), color of clothing (red, blue), among others. We use these attribute annotations on a small dataset of 27k images to train a multi-task CNN, where each head classifies a particular attribute. 
Using the multi-task CNN, we can predict the attributes for new images. 
We denote an attribute vector for an image $I_i$ by $\boldsymbol{a}_i\in\mathbb{R}^A$, where $A$ is the number of attributes. Note that these attributes capture the properties of clothes, not
the identity of people wearing them.

We then learn a set of global styles that capture combinations of
basic attributes by leveraging all the images in the dataset. 
We cluster the feature representation (from the penultimate layer of the CNN) of all the images, using a Gaussian Mixture Model (GMM) with $K=400$ components. 
We denote the style prediction by this model for an image $I_i$ by $s_i\in[1,K]$.  The number of styles is set to yield visually coherent styles, following~\cite{AlHalah-17,Hsiao-17,mall-19,alhalah-20}.

\subsection{Featurizing a Geolocation} \label{ssec:featurization}
We want to characterize locations in a city %
by the fashion sense of their surroundings. 
The fashion in an image  $I_{i}$ with location $l_{i}$ can describe a location $l_{i}$.
However, a location is not described by a single style but a distribution over styles. 
Therefore, we describe the fashion sense of a particular location using the images in its vicinity.
Specifically, to describe a geolocation $\boldsymbol{x}$ we select all images within a %
radius $r$ from that geolocation. 
This formulation lets us capture the distribution of fashion in a surrounding area. 
Additionally, it ensures the fashion distribution changes smoothly from one location to another, and implicitly enforces that nearby locations should belong to the same neighborhood.

Let $T(\boldsymbol{x})$  be the set of images that describes location $\boldsymbol{x}$:
\begin{equation} \label{eq:thresh}
T(\boldsymbol{x}) = \{I_i: ||\boldsymbol{l}_i-\boldsymbol{x}||_2< r\}.
\end{equation}
The images' distribution  over the different styles can describe the location. 
For example, a region near a beach would have a higher distribution over the styles that are unique to a beach (board shorts, tank tops, etc.). 
Therefore, we compute a histogram $\boldsymbol{h_x}\in\mathbb{R^K}$ over the $K$ styles of the images in $T(\boldsymbol{x})$ to describe location
$\boldsymbol{x}$.
Since the sampled images will be biased towards a direction we 
compute an unbiased location for each $\boldsymbol{x}$ (see Supplementary).

There are tradeoffs in choosing the radius $r$.
A small radius would let us create good local features, but would result in a low confidence histogram as there would be very few sampled images. 
A large radius would result in a high confidence histogram, but would capture a larger portion of the map. 
We set the radius so that the ratio of intersection over union of adjacent sampling regions is close to $0.5$.

\subsection{Sampling Locations and Clustering} \label{ssec:clustering}
Next, we use this local featurization at every point to segment the map into different regions.
We cluster regions based on similarities in their histogram descriptors.

Locations are sampled from a uniform 2D grid over the map, at a distance $d$.
If $(w, s)$ is the southwestern location of the city's bounding box and $(W, H)$ is the width and height of the bounding box, a sample location $\boldsymbol(x_{ij})$ is defined by:
\begin{equation} \label{eq:sample}
\boldsymbol{x_{ij}} = (w+di, h+dj) \forall i\in[0, \Bigl\lfloor \frac{W}{d} \Bigr\rfloor, j\in[0, \Bigl\lfloor \frac{H}{d} \Bigr\rfloor].
\end{equation}

We sample images around $\boldsymbol{x_{ij}}$ for all locations $T(\boldsymbol{x_{ij}})$, and obtain the histogram descriptor $\boldsymbol{h_{x_{ij}}}$.
We use K-means to cluster the histogram descriptors to get a label for each location.  %
Since the L1 norm for histograms is 1, we use L1  instead of the standard Euclidean distance when performing the M-step of K-means.

The core of our approach is simple but effective---more effective than other more elaborate variants we explored.
For example, we tried affinity propagation instead of K-means for clustering, but its
neighborhoods 
tended to perform worse %
when adjusted to produce the same number of clusters on our benchmarks.
We also tried an exponential weighting scheme for feature $\boldsymbol{h_x}$, where 
the contribution of an image with location $\boldsymbol{l_i}$ to a histogram for location $\boldsymbol{x}$ is weighted inversely by an exponentiation of $||\boldsymbol{l_i}-\boldsymbol{x}||_2$.
This can be thought of as a softer version of the features in Sec.~\ref{ssec:featurization}.
It performed similarly to hard features, and hence we kept the simple version for evaluation.

In the next three sections we discuss how these discovered neighborhoods can be used for analysis and applications like finding unique neighborhoods of a city, or finding similar and ``analogical" neighborhoods across cities.

\subsection{Finding Unique Neighborhoods}\label{ssec:unique}
Having computed the neighborhoods, we can calculate which neighborhoods have the most unique fashion sense. 
A unique neighborhood is defined in our framework as a neighborhood most distinct from all other neighborhoods in that city. 
Each discovered neighborhood is described by a histogram descriptor $\boldsymbol{h_{n,c}}$ that is created by aggregating images in that neighborhood. 
We use distances between these descriptors to find out the most unique neighborhood, namely, a unique neighborhood has the maximum L1 distance from its most similar neighborhood in the same city:
\begin{equation}
n_{\text{unique}} = \argmax_n{\min_{m \in N, m\ne n}||\boldsymbol{h_{n, c}}-\boldsymbol{h_{m, c}}||_1}.
\end{equation}

We also sort neighborhoods by this distance over all the cities, so that we can rank the most unique neighborhoods with their cities. 

\subsection{Finding Similar Neighborhoods Across Cities}\label{ssec:similarity}
Inspired by applications noted in Sec.~\ref{sec:intro}, we use the discovered neighborhoods to find out similar neighborhoods of two different cities.  
To know which neighborhood of a city is like some other neighborhood in another city, we find the L1 distances between histogram descriptors of all the neighborhoods of all the cities and sort them by this distance.

\begin{figure}[t]
\centering
\includegraphics[width=0.9\linewidth]{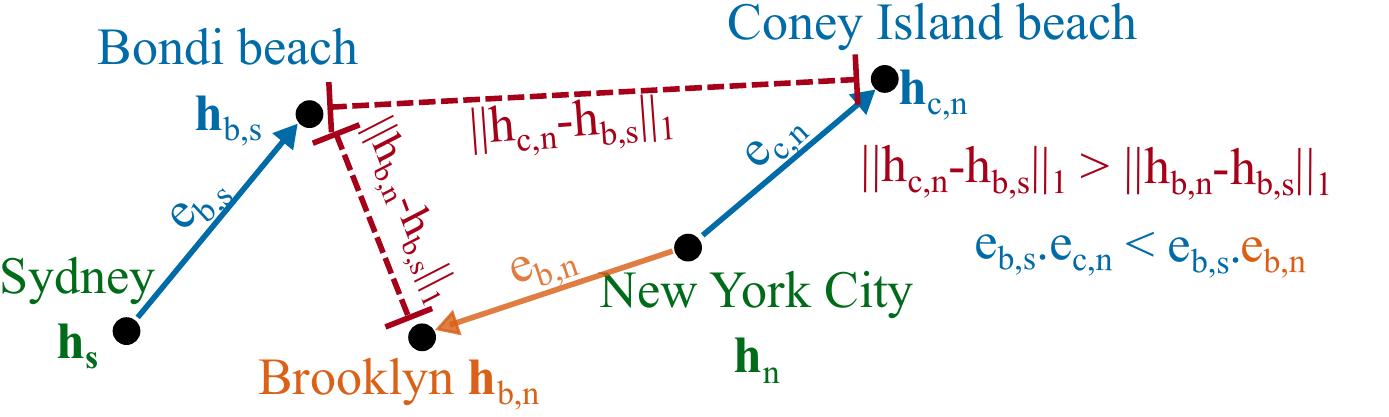}
\caption{\small Illustration of the usefulness of our analogy-inspired encoding.
The two beaches have similar relationships to their city, hence they should be more similar than other neighborhoods---even though their absolute similarity may be low.}
\label{fig:illus}
\vspace*{-0.2in}
\end{figure}

\subsection{Finding Neighborhood Analogies}\label{ssec:analogy}

Finally, we introduce a method to identify \emph{neighborhood analogies}.
The above method is successful in finding similar neighborhoods within cities having similar weather and culture. 
However, if the weather and culture are significantly different for two cities, histogram distances between any of their neighborhoods will be large and hence less meaningful. 
To determine neighborhood analogies, instead of directly measuring distances between neighborhoods, we encode each neighborhood in the context of its city. 
By measuring similarities of neighborhoods using this contextual encoding, we recover analogical pairs of neighborhoods across cities. 
An example of such a pair could be \textit{Bondi beach} : Sydney :: \textit{Coney Island} : NYC, both with popular beaches. Since we are encoding \textit{Bondi beach} with respect to Sydney and \textit{Coney Island} with respect to NYC, the similarity between these two would be a measure of this analogy. 

Each neighborhood has a histogram descriptor $\boldsymbol{h}_{n,c}$, and the city has an aggregate histogram descriptor $\boldsymbol{h}_{c}$. We define the contextual encoding of a neighborhood $n$ with respect to its city $c$ as: 
\begin{equation} e_{n, c} = \text{sgn}(\boldsymbol{h}_{n,c}-\boldsymbol{h}_{c}).
\end{equation}
This encoding contains information about which styles are more popular with respect to other neighborhoods in the city. 
It ignores the magnitudes of relative style popularity and only considers direction, providing invariance to exact style popularities.
We measure the cosine distance between pairs of neighborhoods across cities to find the analogically similar pairs.  
Figure \ref{fig:illus} illustrates how contextual encoding can better capture analogies.
If cities are geographically far apart, there might be a shift in the overall distribution, and a contextual encoding (blue) would produce better results than a non-contextual encoding (red).
Note that if the two cities are similar, both will measure similar quantities.

We implement our method on images from the GeoStyle dataset where each city has 175k images. 
For experiments, we use $r=0.02\degree$ and $d=0.01\degree$.
Exploration of the impact of these hyperparameters is done in the Supplementary.

\vspace*{-0.1in}
\paragraph{Possible dataset biases.}

We employ the GeoStyle dataset~\cite{mall-19} because it is the largest publicly available dataset of its kind and offers a real-world glimpse of what people wear across the world.  While powerful, it has certain limitations.
Images in Instagram (or from any other social media platform) will have sampling biases. 
For example, Instagram is known to be most popular amongst young users.
Given that the dataset is biased to particular age groups, the neighborhoods we discover are also going to be influenced by these age groups.  
The data also focuses on cities rather than rural areas.
Also, a region with tourist attractions is likely to have a higher fraction of photos taken by tourists as compared to non-tourist people in those regions.  Hence, the fraction of tourist vs.~non-tourist images will likely not reflect the true density of tourist vs.~non-tourist populations at a location, 
and we can expect our method to be influenced by tourists and more photogenic places.

\section{Results} \label{sec:results}

We quantitatively evaluate our method's ability to discover underground maps.
Additionally, we look at qualitative results and evaluate the methods with human judgements.  See Supplementary and video for many more examples.

\subsection{Benchmarks}\label{ssec:bench}
To judge our method's ability to segment the city into neighborhoods, %
we need to evaluate it against some notion of ground truth. %
Note that it is impossible to obtain absolute ground truth information of underground maps, %
as there is no single concrete definition of what should be the property constituting underground maps.
Instead we consider multiple approximations of such underground maps for evaluation. 

We create two such benchmarks.
The first benchmark is obtained using an external, publicly crowdsourced platform called HoodMaps\footnote{\url{https://hoodmaps.com/}} and looks at how people from a city perceive different neighborhoods. This benchmark captures subjective impressions. The second benchmark is created using business densities of different types using OpenStreetMap.\footnote{\url{https://www.openstreetmap.org/}}  This benchmark captures objective measures.
Both segment cities into regions based on differences in their properties. %

\begin{figure}[t]
\centering
\includegraphics[width=0.4\linewidth]{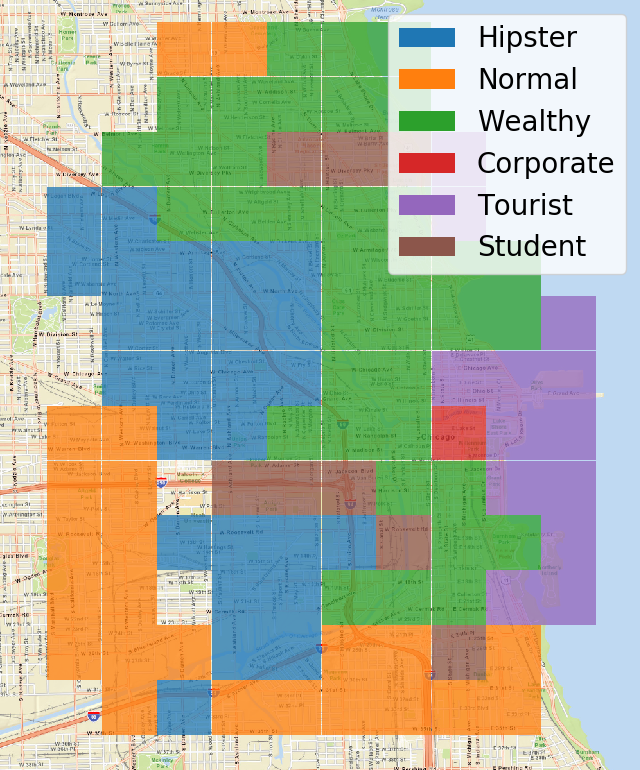}
\includegraphics[width=0.4\linewidth]{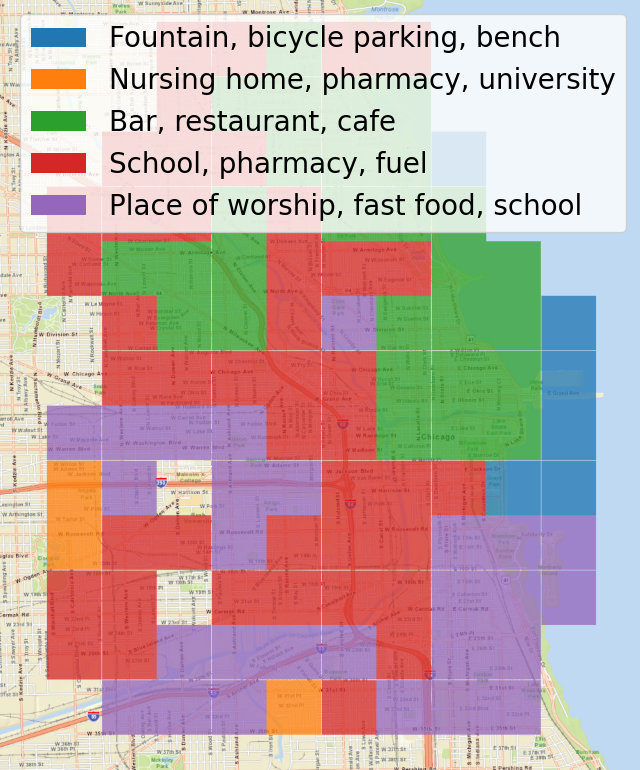}
\vspace*{-0.1in}
\caption{
\small 
We analyze our discovered neighborhoods against two complementary non-visual sources for underground maps. %
\textbf{Left:} HM benchmark labels and neighborhoods dividing Chicago into 6 neighborhoods based on how people perceive them. \textbf{Right:} BD benchmark neighborhoods dividing the city into 5 regions based on its business/amenity distributions. The legend shows 3 businesses that are more frequent in that particular neighborhood.}
\vspace*{-0.1in}
\label{fig:bench}
\end{figure}

\textbf{HoodMap (HM) Benchmark:} We create the HoodMap (HM) Benchmark using information from the crowdsourced public platform called HoodMaps. 
The users on this platform can label regions of a city. %
The service aggregates the votes to provide the majority-voted label for a region. 
Figure \ref{fig:bench} (Left) shows the map of Chicago with the 6 associated labels. %
HoodMaps has a granularity of $0.01^{\degree}$ along both latitude and longitude. 
We collect this information for 37 of the 44 GeoStyle cities with enough data (see Supplementary). %

The annotations in this dataset are based on perceptions of people, and hence they can capture a notion of neighborhoods beyond geographic boundaries. 
For the 37 cities, a region is voted on by more than 70 voters on average, and more than 55\% of voters agree on a single label out of the 6 (chance would be 16\%).
This relatively high number of votes and agreement indicates the labelling is consistent.
However, HoodMaps does come with certain limitations.
The coarse label set was selected by HoodMaps (not us) and is not necessarily complete for all categories of interest.   
The labels may also display certain stereotypes that conflict with the ideal underground map. For example, one of the labels is ``wealthy'', yet our goal is to divide cities on popular activities or interests, \emph{not} socio-economic status.
There is also a possibility of sampling bias amongst people who choose to visit HoodMaps.com and vote for these labels.
In short, while the data is a useful non-visual source for perceived neighborhoods, we also can expect our image-based results to deviate meaningfully from those boundaries, possibly in ways that challenge common stereotypes.

\textbf{Business Distribution (BD) Benchmark:} While the HoodMap benchmark captures how people perceive a neighborhood of a city, it need not capture what activities are present in a neighborhood. 
Therefore, we create the Business Distribution (BD) Benchmark that captures the distribution of different types of businesses across a city. 
We use OpenStreetMap to get geolocations of different business types. 
There are a total of 1,446 different business types and a total of 1.6M businesses/amenities. %
The distribution of the frequency of business types is long-tailed, %
so we consider businesses types with frequency at least 50 (we find 154 such types). 
The distribution of businesses gives a more objective measure of a neighborhood in contrast to HoodMaps.  
For example, a region with a higher distribution of pubs/nightclubs is likely catering to a different crowd than a region with libraries/schools. 
Similarly, a large density of museums indicates a region popular amongst tourists. 
To create regions over maps using business density, we follow our method from Sec.~\ref{sec:method}.
More details are in Supplementary. 
Figure \ref{fig:bench} (Right) shows the BD map of Chicago, along with the amenities/businesses that are more frequent in each region.

In short, the HM and BD benchmarks capture complementary notions of the underground map. 
The former captures how people perceive a neighborhood, whereas the latter captures the activities one can do in a particular neighborhood.

\subsection{Baselines}
As the problem we have introduced is novel, we could not find any prior work as baselines for evaluation.
We evaluate our method against the following set of baselines. %

\noindent\textbf{Random:} is a na\"ive baseline, where every point of interest is assigned a random label.

\noindent\textbf{Proximity:} clusters on geographical proximity instead of style histograms. This baseline uses the uniformly sampled locations (lat.~and long.~pair) as the feature for clustering. 

\noindent\textbf{Proximity+Image Density (PID):} clusters %
on $\boldsymbol{x_{ij}}$
and hence  leverages additional information about the image density at different locations. 
This baseline is stronger than using proximity alone, as image density can tell a lot about a neighborhood, e.g., a residential area is likely to have lower density, unlike a tourist area with lots of photos.

\noindent\textbf{Caption:} clusters image captions. 
Image captions are clustered using aggregated GloVe vectors \cite{pennington-14} for each city and we use histograms over these clusters as features. 
This baseline is similar to our method but uses a non-visual modality instead of visual or fashion-specific cues.

\noindent\textbf{Full image:} sees if the image background provides more useful information about neighborhoods than just looking at the clothing features. 
It uses features from a pre-trained ResNet-18 \cite{he-16} instead of fashion features to create style histograms. It captures global information of people and their background, such as architecture, vegetation, etc.

\noindent\textbf{Administrative boundaries (Admin):}
	represents traditional maps based on government issued boundaries.  
  It uses ordinance maps for the 8 GeoStyle cities for which we could find publicly available data by the city.\footnote{See Supplementary for list of cities. Example for Los Angeles: \cite{LAAdmin}}
  These boundaries are fine-grained, so we greedily merge them based on proximity to match the granularity of the other baselines.

\begin{table*}[t]\small
\centering
      \begin{tabular}{l r r r r r r | r r r r r r} 
        \toprule
         & \multicolumn{6}{c}{All Cities} & \multicolumn{6}{c}{Cities with Administrative Boundaries Available} \\
         & \multicolumn{3}{c}{HM Benchmark} & \multicolumn{3}{c}{BD Benchmark} & \multicolumn{3}{c}{HM Benchmark} & \multicolumn{3}{c}{BD Benchmark}\\
        \textbf{Method} & NMI & Purity & MMIoU & NMI & Purity & MMIoU & NMI & Purity & MMIoU & NMI & Purity & MMIoU\\
        \midrule
        Random & 0.079 & 0.464 & 0.172 & 0.092 & 0.359 & 0.173  & 0.084 & 0.431 & 0.171 & 0.090 & 0.545 & 0.170\\
        Proximity & 0.225 & 0.542 & 0.249 & 0.342 & 0.559 & 0.325 & 0.249 & 0.546 & 0.269 & 0.288 & 0.609 & 0.281\\
        PID & 0.242 & 0.550 & 0.262 & 0.353 & 0.579 & 0.336 & 0.277 & 0.597 & \textbf{0.281} & 0.303 & 0.665 & \textbf{0.294}\\
        Admin & - & - & - & - & - & - & 0.235 & 0.570 & 0.256 & 0.282 & 0.686 & 0.260\\
        Caption & 0.222 & 0.607 & 0.235 & 0.332 & 0.561 & 0.313 & 0.207 & 0.644 & 0.215 & 0.299 & 0.731 & 0.278\\
        Full image & 0.246 & 0.592 & 0.242 & 0.327 & 0.581 & 0.296  & 0.279 & 0.623 & 0.271 & 0.305 & 0.737 & 0.269\\
        Ours & \textbf{0.260} & \textbf{0.635} & \textbf{0.272} & \textbf{0.369}& \textbf{0.597} & \textbf{0.339} & \textbf{0.291} & \textbf{0.652} & \textbf{0.281} & \textbf{0.323}& \textbf{0.742} & 0.281\\
        \bottomrule
      \end{tabular}
      \vspace*{-0.1in}
  \caption{\small Comparison of our method against baselines on both the HM and BD benchmarks. %
  The first six columns show results on all the cities. The last six columns show the results on cities where administrative boundary (Admin) data is available. Our method performs better than all the baselines and all the metrics except for one. %
  \small Our gains over the Admin baseline accentuate how traditional maps (e.g., by city ordinances) are distinct from the underground perceived maps of a city.}
  \vspace*{-0.2in}
  \label{tab:unsupeval}
\end{table*}

\begin{figure}[t]
\centering
\includegraphics[width=\linewidth]{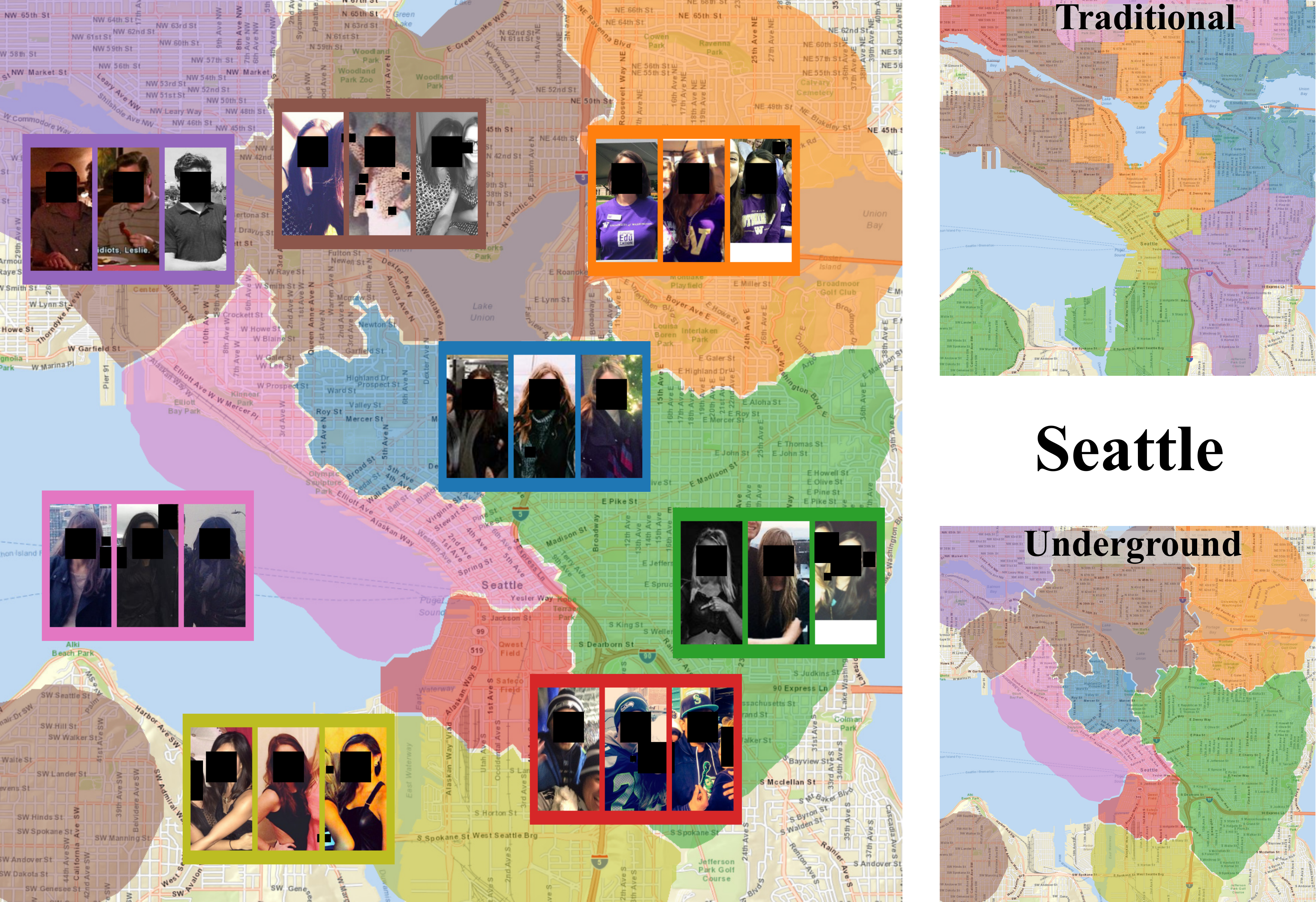}
\vspace*{-0.3in}
\caption{\small Our discovered underground map for Seattle.}
\vspace*{-0.2in}
\label{fig:seattle}
\end{figure}

\subsection{Quantitative Evaluation}

We first evaluate how much neighborhoods discovered by our method align with the neighborhoods of the benchmarks. 
We use 3 unsupervised clustering metrics: (i) Normalized Mutual Information (NMI) captures the mutual information between the produced maps and benchmarks. (ii) Purity captures the maximum precision of the produced maps w.r.t. the benchmarks. (iii) Mean Maximum Intersection over Union (MMIoU) measures the best matching Intersection over Union (IoU). 
We report performance where the number of clusters equals the number of labels in the benchmarks (see Supplementary for sweeps over cluster numbers).
For HoodMaps, this number is 6, and for Business Distribution this number is different for different cities, based on the number of clusters produced by affinity propagation.   %

Table \ref{tab:unsupeval} shows the results.  The left side compares our method to all baselines except Admin aggregated over all 37 cities; the right side compares all methods on the 8 cities for which the baseline Admin is applicable.  Our method outperforms all baselines on both benchmarks on all the metrics for the ``all cities" test, and also outperforms the Admin baseline for every city it is available except for one metric on BD.  Figure~\ref{fig:seattle} shows our discovered map for Seattle, along with the Admin baseline's map (see Supp.~for more).

\textbf{All cities.}
Table~\ref{tab:unsupeval} (left) reveals a few things.
First, the \textit{proximity+image density} baseline performs well and is a very strong baseline.
This shows that the similarity of a region is strongly affected by the image density. %
Second, our performance against the \textit{caption} baseline shows that visual fashion sense provides more useful cues than the manually written captions of the images. 
While \textit{caption} does contain useful information, %
it also contains information not necessarily related to location. 
Third, the \textit{full image} baseline performs worse than our method. 
This shows that fashion specifically is indeed a useful factor for understanding and discovering neighborhoods, beyond the surrounding visual architecture, weather, etc. as captured in the full image. %

\textbf{Cities with administrative boundaries.}
Table~\ref{tab:unsupeval} (right) rescores all baselines and our method for the 8 cities where Admin is possible.  
We see Admin is no better than the proximity baseline. 
This accentuates that the ``traditional" measure of a neighborhood is different from the ``underground" notion.
Figure \ref{fig:fig1} (bottom right) shows the Admin boundaries for New York City for 8 regions:  Admin's manually demarcated neighborhoods cannot find distant similar regions.  
For example, %
whereas we discover regions popular among tourists (red, top right) that are similar to each other, a ``traditional'' neighborhood map of a city will not find such similarities.
Underground maps %
offer significantly different information than traditional maps. 

\textbf{Per-class performance.} 
Next, to understand which underground neighborhood types are best revealed by fashion, we analyze the per-class performance.
Table \ref{tab:perclass} shows the results against the best baseline, PID.\footnote{Here we  measure MMIoU, as purity and NMI are aggregate measures.}  Our method finds it easier to discover tourist and student classes, suggesting it is possible to determine these neighborhoods by looking at changes of particular styles.  Interestingly, the image density itself (PID) reveals cues about hipster and wealthy areas.
Our method finds it difficult to discover corporate neighborhoods.
We believe this is because fewer people are posting images of themselves from a corporate environment.

\begin{table}
\centering
      \begin{tabular}{l r r r r r r r} 
        \toprule
         & Wea. & Hip. & Tou. & Stu. & Nor. & Cor.\\
        \midrule
       PID & 0.297 & 0.281 & 0.247 & 0.223 & 0.256 & 0.171\\
        Ours & 0.293 & 0.289 & 0.265 & 0.243 & 0.258 & 0.166 \\
        \bottomrule

      \end{tabular}\\
\vspace*{-0.1in}
  \caption{\small %
  Per class accuracy on HoodMaps (MMIoU). It is easier to find student and tourist neighborhoods using fashion features, whereas it is difficult to find corporate neighborhoods. Note that the labels were chosen by HoodMaps, and may be associated with stereotypes that we do not aim to discover (see Sec. \ref{ssec:bench}).}
  \label{tab:perclass}
\end{table}

\subsection{Qualitative Results}
Having showed that the discovered neighborhoods  successfully capture underground notions of a city, we now show how we can use these discovered neighborhoods to get useful information about a city and relationships between cities (cf.~Sec.~\ref{ssec:unique} to~\ref{ssec:analogy}).

\begin{figure}
\centering
\includegraphics[width=0.85\linewidth]{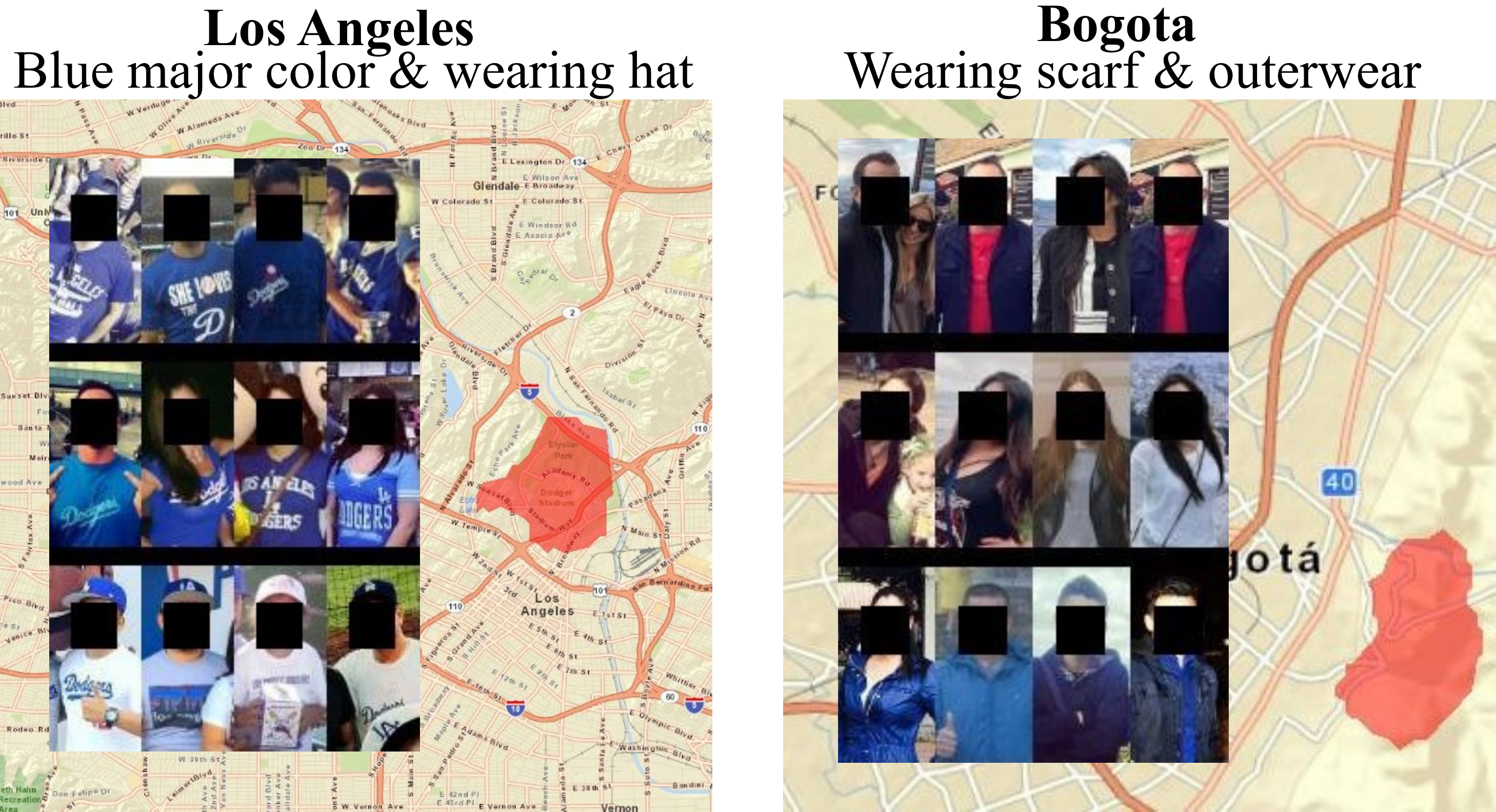}
\caption{\small Two unique neighborhoods. Los Angeles (Left), Bogota
(Right). Each row shows the distinctive style of the neighborhood.}\label{fig:unique}
\vspace*{-0.15in}
\end{figure}

\textbf{Most unique neighborhoods.} 
Figure \ref{fig:unique} shows some of the unique neighborhoods discovered by our method (Sec.~\ref{ssec:unique}).  
First, we find a neighborhood around Dodgers stadium in Los Angeles where a very high number of people can be seen wearing blue colored tops and baseball hats (the team color). 
The second neighborhood in Bogota is popular among tourists for hiking and outdoor activities, where people can be seen wearing glasses and winter clothing. %
See the Supplementary for more unique neighborhoods.

\begin{figure}
\centering
\includegraphics[width=0.85\linewidth]{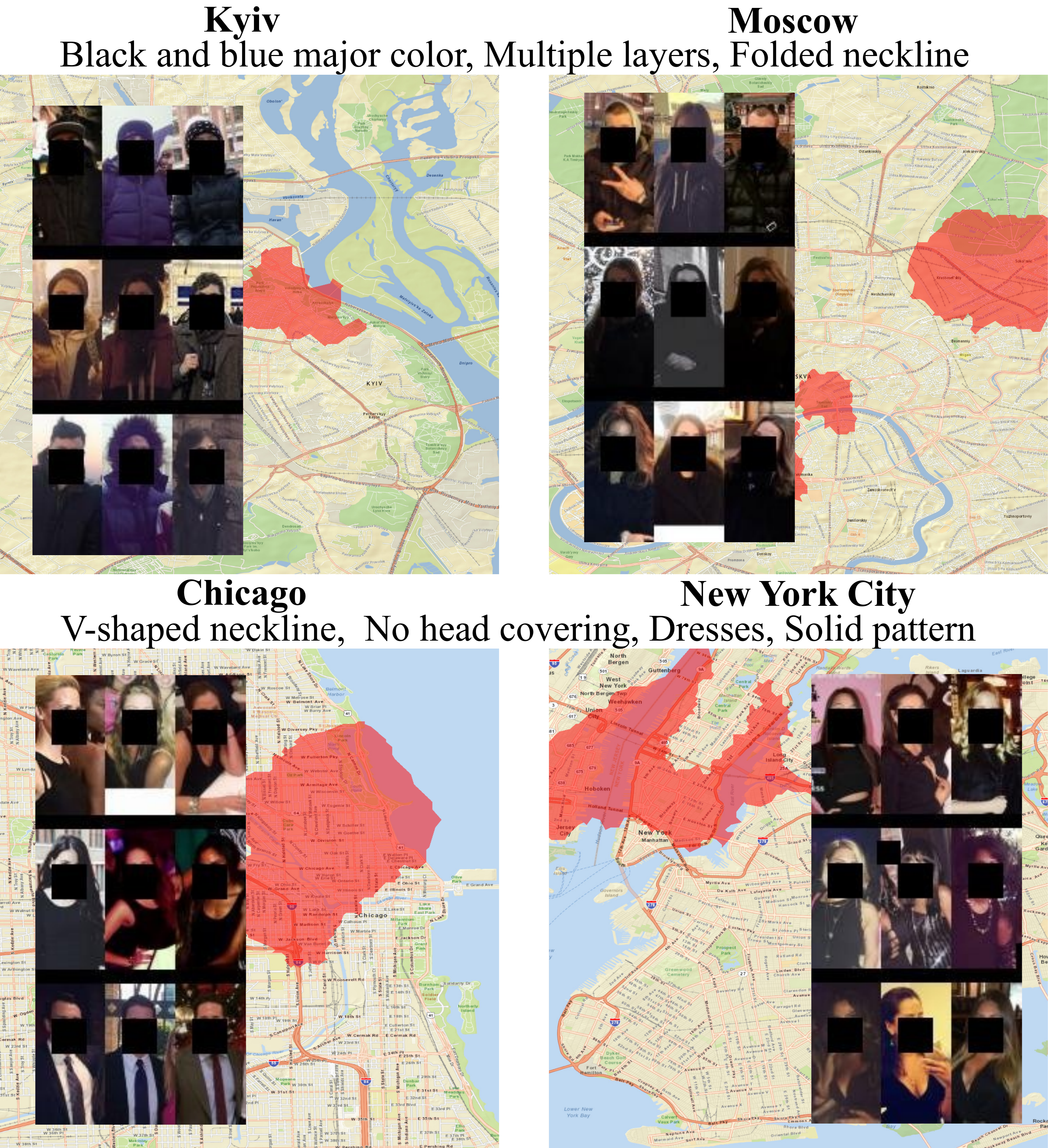}
\vspace*{-0.1in}
\caption{\small Most similar neighborhood between Kyiv-Moscow (top), and Chicago-NYC (bottom). Text on top shows the most popular inferred attributes for the two similar neighborhoods.}\label{fig:simnbr}
\end{figure}

\textbf{Similar neighborhoods across cities.}
Figure \ref{fig:simnbr} shows neighborhoods for Kyiv-Moscow and Chicago-NYC that are similar to each other found using our method in Sec.~\ref{ssec:similarity}. For Kyiv and Moscow, both regions have an unusually high fraction of people in winter clothes. This often signifies tourists in our dataset. As expected these neighborhoods are indeed popular tourist spots for the city of Kyiv (St. Sophia's Cathedral and Independence Square) and Moscow (the Kremlin and Sokolniki Park).
For Chicago and New York City, the similar neighborhoods show a relatively high fraction of people in party wear. Both the neighborhoods are popular places for nightclubs as confirmed by the business density data. (See Supplementary for other examples.)

\begin{figure}
\centering
\includegraphics[width=\linewidth]{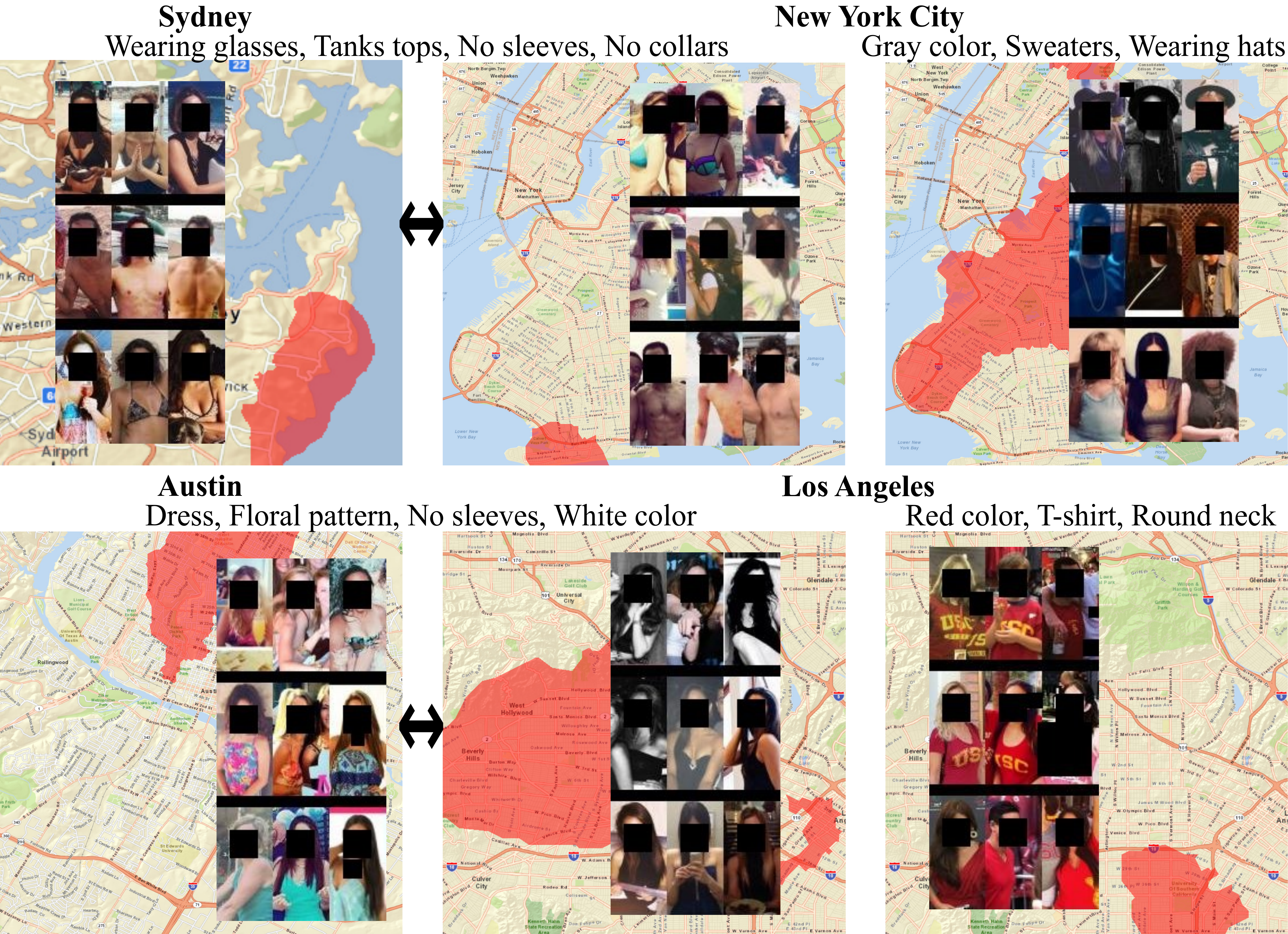}
\vspace*{-0.2in}
\caption{\small Analogy neighborhoods between Sydney and NYC (top) and Austin and Los Angeles (bottom). In each row, the first two images show the analogy obtained with our contextual encoding, and the rightmost image shows the (baseline) similar neighborhood from the second city computed without the contextual encoding.
Text on  top shows the most popular attributes. %
}
\vspace*{-0.1in}
\label{fig:analogy}
\end{figure}

\textbf{Neighborhood analogy.}
Figure \ref{fig:analogy} shows example analogical pairs of neighborhoods found across regions using the proposed contextual encoding (Sec.~\ref{ssec:analogy}). 
Our method is able to discover similar neighborhoods across continents with significantly different weather and culture.
With contextual encoding, we find two neighborhoods with popular beaches in Sydney and New York City (NYC).
Because the histogram shift is too large from Sydney to NYC, if we do not use the proposed contextual encoding an incorrect neighborhood is found (see Figure~\ref{fig:analogy} (top right)).
The difference is further exemplified by the discriminative cluster images shown from the two regions, which highlight beachwear.
Bottom row shows analogical neighborhoods between Austin and Los Angeles. 
Dresses can be seen in the analogical pair as opposed to the neighborhood found without context (right).

\subsection{Experiments with Human Judges}  

A denizen of a city has a deeper understanding of the neighborhoods.  Does %
our technique match what that person would say?
Using Amazon Mechanical Turk to reach ``locals" (see Supplementary), we display a set of images from discovered neighborhoods, and ask the worker to select the point on the map they think the images come from.
Images from neighborhoods of both our method and the strongest baseline (PID) are shown. 
We measure both raw accuracy and the Area Normalized Accuracy (ANA), where per region accuracy is weighted by the inverse of the area of the region. The latter accounts for the fact that the areas of neighborhoods produced by both methods could be unequal, and clicking in the correct region by chance is more probable for the larger region. 

\begin{figure}
\centering
\includegraphics[width=\linewidth]{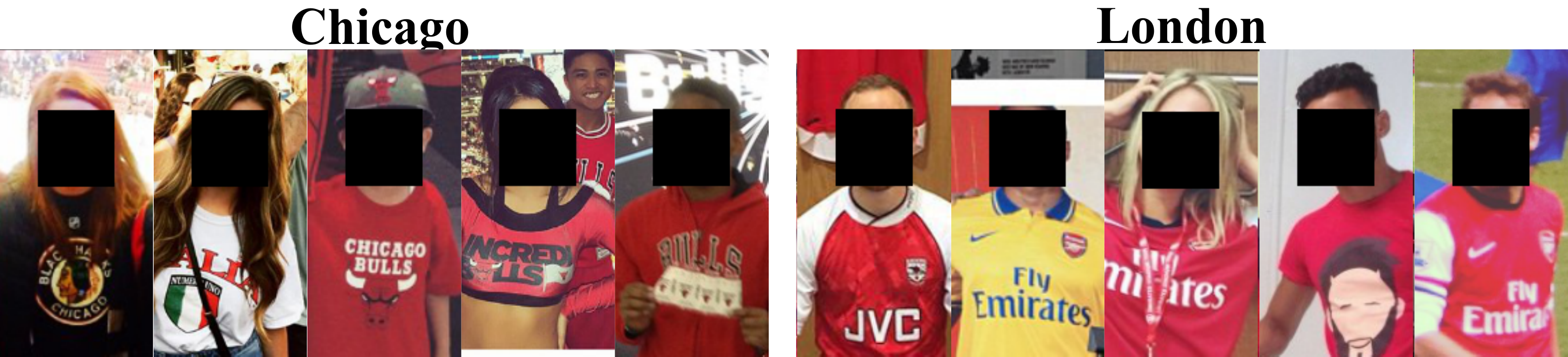}
\vspace*{-0.2in}
\caption{
\small Random samples from two sets of neighborhoods for which all workers confidently point to the correct location on maps.}
\label{fig:human}
\vspace*{-0.2in}
\end{figure}

Table \ref{tab:human} shows the results, accumulated over all cities and judged by at least three workers.
Our method does significantly better than the best PID baseline.
The confidence number shows how confident MTurk users were when marking a region (0: no confidence, 1: very confident). Users were more confident in figuring out the regions when images come from our method.
This result is exciting as it suggests that the neighborhoods discovered automatically by our method are recognizable as coherent by people who know the cities. 
Figure \ref{fig:human} shows examples
for which all workers confidently pointed to the correct place.
Both these neighborhoods contain famous stadiums of the cities' teams (Blackhawks and Arsenal) and are easily identified.

\begin{table}
\centering
      \begin{tabular}{l r r r} 
        \toprule
        \textbf{Method} & Acc. & ANA & Confidence \\
        \midrule
        Proximity+Image Density & 14.28 & 15.00 & 0.57\\
        Ours & \textbf{24.90} & \textbf{27.50} & \textbf{0.61} \\ 
        \bottomrule
      \end{tabular}\\
      \vspace*{-0.1in}
  \caption{\small %
  Human subjects are more frequently able to identify where the neighborhood style comes from when using our neighborhoods. %
  } 
  \label{tab:human}
\vspace*{-0.15in}
\end{table}

\section{Conclusion}
We introduced the concept of underground fashion maps for cities and proposed the first method to create them.
Validating against two new external benchmarks for non-traditional maps, as well as human judges,
we showed that our maps better capture the sense of a neighborhood.  %
We demonstrate various potential uses for such maps, including finding distinctive neighborhoods and analogies across cities.
In future work, we plan to explore how underground maps evolve over time as styles themselves change.

\newpage
{\small
\bibliographystyle{ieee_fullname}
\bibliography{underground}
}

\end{document}